\documentclass{article}

\date{}

\usepackage{amsmath}
\usepackage{bbm}
\usepackage{graphicx}
\usepackage{amsthm}
\usepackage{epstopdf}
\usepackage{subcaption}
\usepackage{wrapfig}
\usepackage{float}
\usepackage[scale=0.85,a4paper]{geometry}
\usepackage{url,hyperref,lineno,microtype}

\def\Authors{Suraj Srinivas, Ravi Kiran Sarvadevabhatla, Konda Reddy Mopuri,  \\ Nikita Prabhu, Srinivas S S Kruthiventi and R. Venkatesh Babu\\ \\ Video Analytics Lab, Indian Institute of Science, Bangalore \\ http://val.serc.iisc.ernet.in/}

\begin{document}
\onecolumn

\title{A Taxonomy of Deep Convolutional Neural Nets for Computer Vision \footnote{This article has been published in Frontiers in Robotics and AI. Please refer to http://goo.gl/6691Bm for the original article.}} 
\author{\Authors} 

\maketitle

\begin{abstract}
Traditional architectures for solving computer vision problems and the degree of success they enjoyed have been heavily reliant on hand-crafted features. However, of late, deep learning techniques have offered a compelling alternative -- that of automatically learning problem-specific features. With this new paradigm, every problem in computer vision is now being re-examined from a deep learning perspective. Therefore, it has become important to understand what kind of deep networks are suitable for a given problem. Although general surveys of this fast-moving paradigm (i.e. deep-networks) exist, a survey specific to computer vision is missing. We specifically consider one form of deep networks widely used in computer vision - convolutional neural networks (CNNs). We start with ``AlexNet'' as our base CNN and then examine the broad variations proposed over time to suit different applications. We hope that our recipe-style survey will serve as a guide, particularly for novice practitioners intending to use deep-learning techniques for computer vision.

\end{abstract}

\section{Introduction}
Computer vision problems like image classification and object detection have traditionally been approached using hand-engineered features like SIFT by \cite{lowe2004distinctive} and HoG by \cite{dalal2005histograms}. Representations based on the Bag-of-visual-words descriptor (see \cite{yang2007evaluating}), in particular, enjoyed success in image classification. These were usually followed by learning algorithms like Support Vector Machines (SVMs). As a result, the performance of these algorithms relied crucially on the features used. This meant that progress in computer vision was based on hand-engineering better sets of features. With time, these features started becoming more and more complex - resulting in a difficulty with coming up better, more complex features. From the perspective of the computer vision practitioner, there were two steps to be followed: feature design and learning algorithm design, both of which were largely independent.

Meanwhile, some researchers in the machine learning community had been working on learning models which incorporated learning of features from raw images. These models typically consisted of multiple layers of non-linearity. This property was considered to be very important - and this lead to the development of the first deep learning models. Early examples like Restricted Boltzmann Machines (\cite{hinton2002training}) , Deep Belief Networks (\cite{hinton2006fast}) and Stacked Autoencoders (\cite{vincent2010stacked}) showed promise on small datasets. The primary idea behind these works was to leverage the vast amount of unlabelled data to train models. This was called the `unsupervised pre-training' stage. It was believed that these `pre-trained' models would serve as a good initialization for further supervised tasks such as image classification. Efforts to scale these algorithms on larger datasets culminated in 2012 during the ILSVRC competition (see \cite{russakovsky2014imagenet}), which involved, among other things - the task of classifying an image into one of thousand categories. For the first time, a Convolutional Neural Network (CNN) based deep learnt model by \cite{krizhevsky2012imagenet} brought down the error rate on that task by half, beating traditional hand-engineered approaches. Surprisingly, this could be achieved by performing end-to-end supervised training, without the need for unsupervised pre-training. Over the next couple of years, `Imagenet classification using deep neural networks' by \cite{krizhevsky2012imagenet} became one of the most influential papers in computer vision. Convolutional Neural Networks, a particular form of deep learning models, have since been widely adopted by the vision community. In particular, the network trained by Alex Krizhevsky, popularly called ``AlexNet'' has been used and modified for various vision problems. Hence, in this article, we primarily discuss CNNs, as they are more relevant to the vision community. With the plethora of deep convolutional networks that exist for solving different tasks, we feel the time is right to summarize CNNs for a survey. This article can also serve as a guide for beginning practitioners in deep learning/computer vision. 

The paper is organized as follows. We first develop the general principles behind CNNs (Section 2), and then discuss various modifications to suit different problems (Section 3). Finally, we discuss some open problems (Section 4) and directions for further research.

\section{Introduction to Convolutional Neural Networks}

The idea of a Convolutional Neural Network (CNN) is not new. This model had been shown  to work well for hand-written digit recognition by \cite{lecun1998gradient}. However, due to the inability of these networks to scale to much larger images, they slowly fell out of favour. This was largely due to memory and hardware constraints, and the unavailability of large amounts of training data. With increase in computational power thanks to wide availability of GPUs, and the introduction of large scale datasets like the ImageNet (see \cite{russakovsky2014imagenet}) and the MIT Places dataset (see \cite{zhou2014learning}), it was possible to train larger, more complex models. This was first shown by the popular \textit{AlexNet} model which was discussed earlier. This largely kick-started the usage of deep networks in computer vision. 

\subsection{Building Blocks of CNNs}
\begin{figure}
\centering
\includegraphics[height=7cm,trim={0cm 30cm 0cm 0cm},clip]{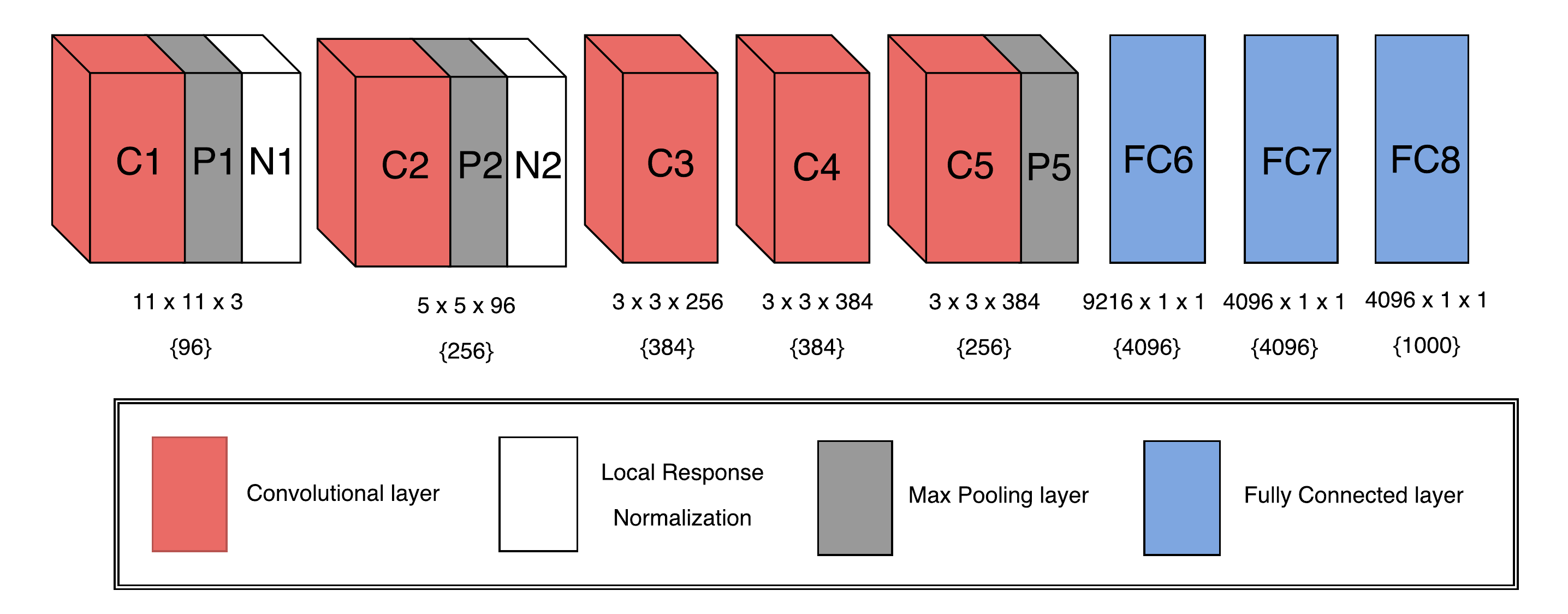}
\caption{An illustration of the weights in the AlexNet model. Note that after every layer, there is an implicit ReLU non-linearity. The number inside curly braces represents the number of filters with dimensions mentioned above it.}
\label{fig:alexnet}
\end{figure}

In this section, we shall look at the basic building blocks of CNNs in general. This assumes that the reader is familiar with traditional neural networks, which we shall call ``fully connected layers'' in this article. Figure \ref{fig:alexnet} shows a representation of the weights in the \textit{AlexNet} model. While the first five layers are convolutional, the last three are fully connected layers. 

\subsubsection{Why convolutions?}

Using traditional neural networks for real-world image classification is impractical for the following reason: Consider a 2D image of size $200 \times 200$ for which we would have have $40,000$ input nodes. If the hidden layer has $20,000$ nodes, the size of the matrix of input weights would be $40,000 \times 20,000 = 800$ Million. This is just for the first layer -- as we increase the number of layers, this number increases even more rapidly. Besides, vectorizing an image completely ignores the complex 2D spatial structure of the image. How do we build a system that overcomes both these disadvantages?

\indent One way is to use 2D convolutions instead of matrix multiplications. Learning a set of convolutional filters (each of $11 \times 11$, say) is much more tractable than learning a large matrix ($40000 \times 20000$). 2D convolutions also naturally take the 2D structure of images into account. Alternately, convolutions can also be thought of as regular neural networks with two constraints (See \cite{bishop2006pattern}):
\begin{itemize}
\item Local connectivity: This comes from the fact that we use a convolutional filter with dimensions much smaller than the image it operates on. This contrasts with the \textit{global} connectivity paradigm typically relevant to vectorized images.
\item Weight sharing: This comes from the fact that we perform convolutions, i.e. we apply the same filter across the image. This means that we use the same \textit{local} filters on many locations in the image. In other words, the weights between all these filters are shared.
\end{itemize}

There is also evidence from visual neuroscience for similar computations within the human brain. \cite{hubel1962receptive} found two types of cells in the primary visual cortex - the simple cells and the complex cells. The simple cell responded primarily to oriented edges and gratings - which are reminiscent of Gabor filters, a special class of convolutional filters. The complex cells  were also sensitive to these egdes and grating. However, they exhibited spatial invariance as well. This motivated the Neocognitron model by \cite{fukushima1980neocognitron}, which proposed the learning of convolutional filters in an artificial neural network. This model is said to have inspired convolutional networks, which are analogous to the simple cells mentioned above.

\indent In practical CNNs however, the convolution operations are not applied in the traditional sense wherein the filter shifts one position to the right after each multiplication. Instead, it is common to use larger shifts (commonly referred to as stride). This is equivalent to performing image down-sampling after regular convolution. 

\indent If we wish to train these networks on RGB images, one would need to learn multiple \textit{multi-channel} filters. In the representation in Figure \ref{fig:alexnet}, the numbers $11 \times 11 \times 3$, along with $\{96\}$ below \textbf{C1} indicates that there are $96$ filters in the first layers, each of spatial dimension of $11 \times 11$, with one for each of the $3$ RGB channels.

We note that this paradigm of convolution like operations (location independent feature-detectors) is not entirely suitable for registered images. As an example, images of faces require different feature-detectors at different spatial locations. To account for this, \cite{taigman2014deepface} consider only locally-connected networks with no weight-sharing. Thus, the choice of layer connectivity depends on the underlying type of problem.

\subsubsection{Max-Pooling}
The Neocognitron model inspired the modelling of simple cells as convolutions. Continuing in the same vein, the complex cells can be modelled as a max-pooling operation. This operation can be thought of as a \textit{max filter}, where each $n \times n$ region is replaced with it's max value. 
This operation serves two purposes:
\begin{enumerate}
\item It picks out the highest activation in a local region, thereby providing a small degree of spatial invariance. This is analogous to the operation of complex cells.
\item It reduces the size of the activation for the next layer by a factor of $n^2$. With a smaller activation size, we need a smaller number of parameters to be learnt in the later layers.
\end{enumerate}

\subsubsection{Non-linearity}

Deep networks usually consist of convolutions followed by a non-linear operation after each layer. This is necessary because cascading linear systems (like convolutions) is another linear system. Non-linearities between layers ensure that the model is more expressive than a linear model. 

In theory, no non-linearity has more expressive power than any other, as long as they are continuous, bounded and monotonically increasing (see \cite{hornik1991approximation}). Traditional feedforward neural networks used the sigmoid ($\sigma(x) = \frac{1}{1+e^{-x}} $) or the tanh (tanh($x$) $= \frac{e^x - e^{-x}}{e^x + e^{-x}}$) non-linearities. However, modern convolutional networks use the ReLU (ReLU($x$) $= max(0,x)$) non-linearity. CNNs with this non-linearity have been found to train faster, as shown by \cite{nair2010rectified}.

\indent Recently, \cite{maas2013rectifier} introduced a new kind of non-linearity, called the leaky-ReLU. It was defined as Leaky-ReLU($x$) $= max(0,x) + \alpha min(0,x)$, where $\alpha$ is a pre-determined parameter. \cite{he2015delving} improved on this by suggesting that the $\alpha$ parameter also be learnt, leading to a much richer model.

\subsection{Depth}
The Universal Approximation theorem by \cite{hornik1991approximation} states that a neural network with a single hidden layer is sufficient to model any continuous function. However, \cite{bengio2009learning} showed that such networks need an exponentially large number of neurons when compared to a neural network with many hidden layers. Recently, \cite{romero2014fitnets}, and \cite{ba2014deep} explicitly showed that a deeper neural network can be trained to perform much better than a comparatively shallow network.

\indent Although the motivation for creating deeper networks was clear, for a long time researchers did not have an algorithm that could efficiently train neural networks with more than 3 layers. With the introduction of greedy layerwise pre-training by \cite{hinton2006fast}, researchers were able to train much deeper networks. This played a major role in bringing the so-called \textit{Deep Learning} systems into mainstream machine learning. Modern deep networks such as AlexNet have 7 layers. More recent networks like VGGnet by \cite{simonyan2014very} and GoogleNet by \cite{szegedy2014going} have 19 and 22 layers respectively were shown to perform much better than AlexNet.

\subsection{Learning algorithm}
A powerful, expressive model is of no use without an algorithm to learn the model's parameters efficiently. The greedy layerwise pre-training approaches in the pre-AlexNet era attempted to create such an efficient algorithm. However, for computer vision tasks, it turned out that a simpler supervised training procedure was enough to learn a powerful model. 

\indent Learning is generally performed by minimization of certain loss functions. Tasks based on classification use the softmax loss function or the sigmoid cross entropy function, while those involving regression use the euclidean error function. In the example of Figure \ref{fig:alexnet}, the output of the \texttt{FC8} layer is trained to represent one of thousand classes of the dataset.

\subsubsection{Gradient-based optimization}
\indent Neural networks are generally trained using the backpropogation algorithm (see \cite{rumelhart1988learning}), which uses the chain rule to speed up the computation of the gradient for the gradient descent (GD) algorithm. However, for datasets with thousands (or more) of data points, using GD is impractical. In such cases, an approximation called the Stochastic Gradient Descent (SGD) is used. It has been found that training using SGD generalizes much better than training using GD. However, one disadvantage is that SGD is very slow to converge. To counteract this, SGD is typically used with a mini-batch, where the mini-batch typically contains a small number of data-points ($\sim 100$). 

\indent Momentum (see \cite{polyak1964some}) belongs to a family of methods that aim to speed the convergence of SGD. This is largely used in practice to train deep networks, and is often considered as an essential component. Other extensions like Adagrad by \cite{duchi2011adaptive}, Nesterov's accelerated GD by \cite{nesterov1983method} , Adadelta by \cite{zeiler2012adadelta} and Adam by \cite{kingma2014adam} are known to work equally well, if not better than vanilla momentum in certain cases. For detailed discussion on how these methods work, the reader is encouraged to read \cite{sutskever2013importance}.

\subsubsection{Dropout}
When training a network with a large number of parameters, an effective regularization mechanism is essential to combat overfitting. Usual approaches such as $\ell_1$ or $\ell_2$ regularization on the weights of the neural net have been found to be insufficient in this aspect. Dropout is a powerful regularization method introduced by \cite{hinton2012improving} which has been shown to work well for large neural nets. 
\indent To use dropout, we \textit{randomly} drop neurons with a probability $p$ during training. As a result, only a random subset of neurons are trained in a single iteration of SGD. At test time, we use all neurons, however we simply multiply the activation of each neuron with $p$ to account for the scaling. \cite{hinton2012improving} showed that this procedure was equivalent to training a large ensemble of neural nets with shared parameters, and then using their geometric mean to obtain a single prediction.

\indent Many extensions to dropout like DropConnect by \cite{wan2013regularization} and Fast Dropout by \cite{wang2013fast} have been shown to work better in certain cases. Maxout by \cite{goodfellow2013maxout} is a non-linearity that improves performance of a network which uses dropout. 

\subsection{Tricks to increase performance}
While the techniques and components described above are theoretically well-grounded, certain \textit{tricks} are crucial to obtaining \textit{state-of-the-art} performance. 

It is well known that machine learning models perform better in the presence of more data. Data augmentation is a process by which some geometric transforms are applied to training data to increase their number. Some examples of commonly used geometric transforms include random cropping, RGB jittering, image flipping and small rotations. It has been found that using augmented data typically boosts performance by about 3\%  (see \cite{ken2014return}). 

Also well-known is the fact that an ensemble of models perform better than one. Hence, it is the commonplace to train several CNNs and average their predictions at test time. Using ensembles has been found to typically boost accuracy by 1-2\% (see \cite{simonyan2014very} and \cite{szegedy2014going}) .

\subsection{Putting it all together: AlexNet}

The building blocks discussed above largely describe AlexNet as a whole. As shown in Figure \ref{fig:alexnet}, only layers 1,2 and 5 contain max-pooling, while dropout is only applied to the last two fully connected layers as they contain the most number of parameters. Layers 1 and 2 also contain Local Response Normalization, which has not been discussed as \cite{ken2014return} showed that its absence does not impact performance.

\indent This network was trained on the ILSVRC 2012 training data, which contained 1.2 million training images belonging to 1000 classes. This was trained on 2 GPUs over the course of one month. The same network can be trained today in little under a week using more powerful GPUs (see \cite{ken2014return}). The hyper-parameters of the learning algorithms like learning rate, momentum, dropout and weight decay were hand tuned.
\indent It is also interesting to note the trends in the nature of features learnt at different layers. The earlier layers tend to learn gabor-like oriented edges and blob-like features, followed by layers that seem to learn more higher order features like shapes. The very last layers seem to learn semantic attributes such as eyes or wheels, which are crucial parts in several categories. A method to visualize these was provided by \cite{zeiler2014visualizing}.

\subsection{Using Pre-trained CNNs}
One of the main reasons for the success of the AlexNet model was that it was possible to directly use the pre-trained model to do various other tasks which it was not originally intended for. It became remarkably easy to download a learnt model, and then tweak it slightly to suit the application at hand. We describe two such ways to use models in this manner.

\subsubsection{Fine-tuning}
Given a model trained for image classification, how does one modify it to perform a different (but related) task? The answer is to just use the trained weights as an initialization and run SGD again for this new task. Typically, one uses a learning rate much lower than what was used for learning the original net. If the new task is very similar to the task of image classification (with similar categories), then one need not re-learn a lot of layers. The earlier layers can be fixed and only the later, more semantic layers need to be re-learnt. However, if the new task is very different, one ought to either re-learn all layers, or learn everything from scratch. The number of layers to re-learn also depends on the number of data points available for training the new task. The more the data, the higher is the number of layers that can be re-learnt. The reader is urged to refer to \cite{yosinski2014transferable} for more thorough guidelines.

\subsubsection{CNN activations as features}
As remarked earlier, the later layers in AlexNet seem to learn visually semantic attributes. These intermediate representations are crucial in performing 1000-way classification. Since these represent a wide variety of classes, one can use the FC7 activation of an image as a generic feature descriptor. These features have been found to be better than hand-crafted features like SIFT or HoG for various computer vision tasks. 

\indent \cite{donahue2013decaf} first introduced the idea of using CNN activations as features and performed tests to determine their suitability for various tasks. \cite{babenko2014neural} proposed to use the activations of fully-connected layers for image retrieval, which they dubbed ``Neural Codes''. \cite{razavian2014cnn} used these activations for various tasks and concluded that off-the-shelf CNN features can serve as a hard-to-beat baseline for many tasks. \cite{hariharan2014hypercolumns} used activations \textit{across} layers as a feature. Specifically, they look at the activations produced by a single image pixels across the network and pool them together. They were found to be useful for fine-grained tasks such as keypoint localization.  

\subsection{Improving AlexNet}

The performance of AlexNet motivated a number of CNN-based approaches, all aimed at a performance improvement over and above that of AlexNet's. Just as AlexNet was the winner for ILSVRC challenge in 2012, a CNN-based net \texttt{Overfeat} by \cite{sermanet2013overfeat} was the top-performer at ILSVRC-2013. Their key insight was that  training a convolutional network to simultaneously classify, locate and detect objects in images can boost the classification accuracy and the detection and localization accuracy of all tasks. Given its multi-task learning paradigm, we discuss \texttt{Overfeat} when we discuss hybrid CNNs and multi-task learning in Section 3.5. 

GoogleNet by \cite{szegedy2014going}, the top-performer at ILSVRC-2014, established that very deep networks can translate to significant gains in classification performance. Since naively increasing the number of layers results in a large number of parameters, the authors employ a number of ``design tricks''. One such trick is to have a trivial $1 \times 1$ convolutional layer after a regular convolutional layer. This has the net effect of not only reducing the number of parameters, but also results in CNNs with more expressive power. This design trick is laid out in better detail in the work of ~\cite{szegedy2014going} where the authors show that having one or more $1 \times 1$ convolutional layers is akin to having a multi-layer perceptron network processing the outputs of the convolutional layer that precedes it. Another trick that the authors utilize is to involve inner layers of the network in the computation of the objective function instead of the typical final softmax layer (as in AlexNet). The authors attribute scale invariance as the reason behind this design decision.

VGG-19 and its variants by \cite{simonyan2014very} is another example of a high-performing CNN where the deeper-is-better philosophy is applied in the net design. An interesting feature of VGG design is that it forgoes larger sized convolutional filters for stacks of smaller sized filters. These smaller sized filters tend to be chosen so that they contain approximately the same number of parameters as the larger filters they supposedly replace. The net effect of this design decision is efficiency and regularization-like effect on parameters due to the smaller size of the filters involved.

\section{CNN Flavours}

\subsection{Region-based CNNs}

Most CNNs trained for image recognition are trained using a dataset of images containing a single object. At test time, even in case of multiple objects, the CNN may still predict a single class. This inherent problem with the design of the CNNs is not restricted to image classification alone. For example, the problem of object detection and localization requires not only classifying the image but also estimating the class and precise location of the object(s) present in the image. Object detection is challenging since we potentially want to detect multiple objects with varying sizes within a single image. It generally requires processing the image patch-wise, looking for the presence of objects. Neural nets have been employed in this way for detecting specific objects like faces in \cite{vaillant1994original} and \cite{rowley-98} and for pedestrians by \cite{lecun-cvpr-13}.

Meanwhile, detecting a set of object-like regions in a given image - also called region proposals or object proposals - has gained a lot of attention (see \cite{selective-search}). These region proposals are class agnostic and reduce the overhead incurred by the traditional exhaustive sliding window approach. These region proposal algorithms operate at low level and output hundreds of object like image patches at multiple scales. In order to employ a classification net towards the task of object localization, image patches of different scales have to be searched one at a time.

\begin{figure}
\centering
\includegraphics[width=15cm,trim={0cm 0cm 0cm 2.8cm},clip]{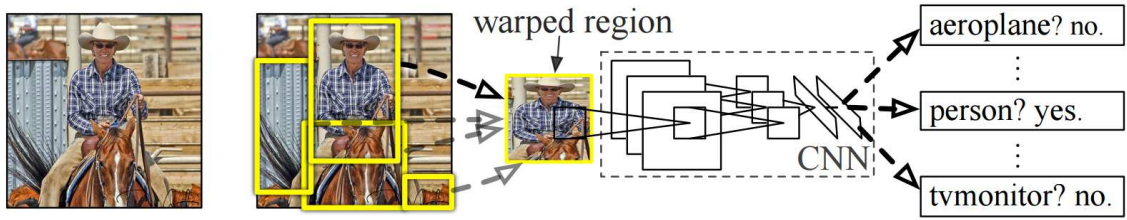}
\caption{Object detection system of \cite{girshick2014rich} using deep features extracted from image regions}
\label{fig:rcnn}
\end{figure}

Recent work by \cite{girshick2014rich} attempt to solve the object localization problem using a set of region proposals. During test time, the method generates around $2000$ category independent region proposals using selective search by \cite{selective-search} from the test image. They employ a simple affine image warping to feed each of these proposals to a CNN trained for classification. The CNN then describes each of these regions with a fixed size high level semantic feature. Finally, a set of category specific linear SVMs classify each region, as shown in Figure \ref{fig:rcnn}. This method achieved the best detection results on the PASCAL VOC 2012 dataset. As this method uses image regions followed by a CNN, it is dubbed R-CNN (Region-based CNN).

A series of works adapted the R-CNN approach to extract richer set of features at patch or region level to solve a wide range of target applications in vision. However, CNN representations lack robustness to geometric transformations restricting their usage. \cite{mopcnn} show empirical evidence that the global CNN features are sensitive to general transformations such as translation, rotation and scaling. In their experiments, they report that this inability of global CNN features translates directly into a loss in the classification accuracy. They proposed a simple technique to pool the CNN activations extracted from the local image patches. The method extracts image patches in an exhaustive sliding-window manner at different scales and describes each of them using a CNN. The resulting dense CNN features are pooled using VLAD (see \cite{vlad-pami-11}) in order to result in a representation which incorporates spatial as well as semantic information.

Instead of considering the image patches at exhaustive scales and image locations, \cite{mopuri-cvprw-15} utilize the objectness prior to automatically extract the image patches at different scales. They build a more robust image representation by aggregating the individual CNN features from the patches for an image search application.

\cite{wei-corr-14}  extended the capability of a CNN that is trained to output a single label into predicting multiple labels. They consider an arbitrary number of region proposals in an image and share a common CNN across all of them in order to obtain individual predictions. Finally, they employ a simple pooling technique to produce the final multi-label prediction.

\subsection{Fully Convolutional Networks}

The success of Convolutional Neural Networks in the tasks of image classification ~(see \cite{krizhevsky2012imagenet, szegedy2014going}) and object detection ~(see \cite{girshick2014rich}) has inspired researchers to use deep networks for more challenging recognition problems like semantic object segmentation and scene parsing. Unlike image classification, semantic segmentation and scene parsing are problems of structured prediction where every pixel in the image grid needs to be assigned a label of the class to which it belongs (e.g., road, sofa, table etc.). This problem of per-pixel classification has been traditionally approached  by generating region-level (e.g. superpixel) hand crafted features and classifying them using a Support Vector Machine (SVM) into one of the possible classes. 

Doing away with these engineered features, \cite{farabet2013learning} used hierarchical learned features from a convolutional neural net for scene parsing. Their approach comprised of densely computing multi-scale CNN features for each pixel and aggregating them over image regions upon which they are classified. However, their method still required the post-processing step of generating over-segmented regions, like superpixels, for obtaining the final segmentation result. Additionally, the CNNs used for multi-scale feature learning were not very deep with only three convolution layers.

\begin{figure}
\centering
\includegraphics[width = \linewidth]{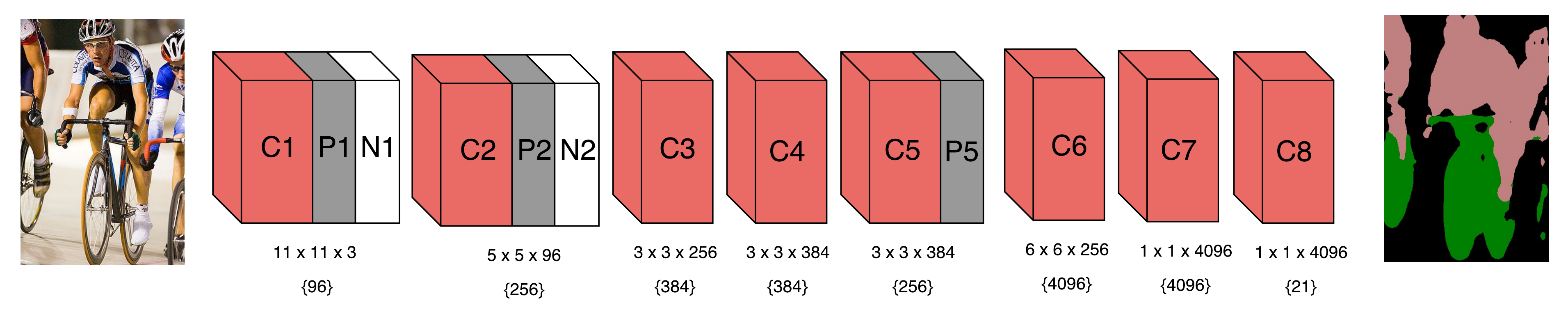}
\caption{Fully Convolutional Net: AlexNet modified to be fully convolutional for performing semantic object segmentation on PASCAL VOC 2012 dataset with 21 classes}
\label{fig:FCNAlexNet}
\end{figure}

Later, \cite{long2015fully} proposed a fully convolutional network architecture for learning per-pixel tasks, like semantic segmentation, in an end-to-end manner. This is shown in Figure \ref{fig:FCNAlexNet}. Each layer in the fully convolutional net (FullConvNet) performs a location invariant operation i.e., a spatial shift of values in the input to the layer will only result in an equivalent scaled spatial shift in its output while keeping the values nearly intact. This property of translational invariance holds true for the convolutional and maxpool layers which form the major building blocks of a FullConvNet. Further, these layers have an output-centred, fixed-size receptive field on its input blob. These properties of the layers of FullConv Net allow it to retain the spatial structure present in the input image in all of its intermediate and final outputs. 

Unlike CNNs used for image classification, a FullConvNet does not contain any densely connected/inner product layers as they are not translation invariant. The restriction on the size of input image to a classification CNN (e.g., 227x227 for AlexNet~\cite{krizhevsky2012imagenet}, 224x224 for VGG~\cite{simonyan2014very}) is imposed due to the constraint on the input size to its inner product layers. Since a FullConvNet does not have any of these inner product layers, it can essentially operate on input images of any arbitrary size.

During the design of CNN architectures, one has to make a trade-off between the number of channels and the spatial dimensions for the data as it passes through each layer. Generally, the number of channels in the data are made to increase progressively while bringing down its spatial resolution, by introducing stride in the convolution and max-pool layers of the net. This is found to be an effective strategy for generating richer semantic representations in a hierarchical manner. While this method enables the net to recognize complex patterns in the data, it also diminishes the spatial resolution of the data blob progressively after each layer. While this is not a major concern for classification nets which require only a single label for the entire image, this results in per-pixel prediction only at a sub-sampled resolution in case of FullConvNets. For tackling this problem, \cite{long2015fully} have proposed a deconvolution layer which brings back the spatial resolution from the sub-sampled output through a learned upsampling operation. This upsampling operation is performed at intermediate layers of various spatial dimensions and are concatenated to obtain pixel-level features at the original resolution.

On the other hand, \cite{chen2014semantic} adopted a more simplistic approach for maintaining resolution by removing the stride in the layers of FullConvNet, wherever possible. Following this, the FullConvNet predicted output is modeled as a unary term for Conditional Random Field (CRF) constructed over the image grid at its original resolution. With labelling smoothness constraint enforced through pair-wise terms, the per-pixel classification task is modeled as a CRF inference problem. While this post-processing of FullConvNet's coarse labelling using CRF has been shown to be effective for pixel-accurate segmentation, \cite{zheng2015conditional} have proposed a better approach where the CRF constructed on image is modeled as a Recurrent Neural Network (RNN). By modeling the CRF as an RNN, it can be integrated as a part of any Deep Convolutinal Net making the system efficient at both semantic feature extraction and fine-grained structure prediction. This enables the end-to-end training of the entire FullConvNet + RNN system using the stochastic gradient descent (SGD) algorithm to obtain fine pixel-level segmentation.

Visual Saliency Prediction is another important problem considered by researchers. This task involves predicting the salient regions of an image given by human eye fixations. Works by \cite{vig2014large} and \cite{liu2015predicting} proposed CNN-based approaches for estimating the saliency score for constituent image patches using deep features. As a result, they did not use a FullConvNet architecture. In contrast, \cite{kruthiventi2015deepfix} proposed a fully convolutional architecture - DeepFix which learnt to predict saliency for the entire image in an end-to-end fashion and attained a superior performance. Their network characterized the multi-scale aspects of the image using inception blocks and captured the global context using convolutional layers with large receptive fields. Another work, by \cite{li2015deepsaliency}, proposed a multi-task FullConvNet architecture - DeepSaliency for joint saliency detection and semantic object segmentation. Their work showed that learning features collaboratively for two related prediction tasks can boost overall performance.


\subsection{Multi-modal networks}

The success of CNNs on standard RGB vision tasks is naturally extended to works on other perception modalities like RGB-D and motion information in the videos. Recently, there has been an increasing evidence for the successful adaptation of the CNNs to learn efficient representations from the depth images. \cite{socher-nips-12} exploited the information from color and depth modalities for addressing the problem of classification. In their approach, a single layer of CNN extracts low level features from both the RGB and depth images separately. These low level features from each modality are given to a set of RNNs for embedding into a lower dimension. Concatenation of the resulting features forms the input to the final soft-max layer. The work by \cite{couprie-corr-13} extended the CNN method of \cite{farabet-pami-13} to label the indoor scenes by treating depth information as an additional channel to the existing RGB data. Similarly, \cite{wang-eccv-14} adapt an unsupervised feature learning approach to scene labeling using RGB-D input with four channels. \cite{gupta-eccv-14} proposed an encoding for the depth images that allows CNNs to learn stronger features than from the depth image alone. They encode depth image into three channels at each pixel: horizontal disparity, height above ground, and the angle the pixel's local surface normal makes with the inferred gravity direction. Their approach for object detection and segmentation processes RGB and the encoded depth channels separately. The learned features are fused by concatenating and further fed into a SVM. 

Similarly, one can think of extending these works for video representation and understanding. When compared to still images, videos provide important additional information in the form of motion. However, majority of the early works that attempted to extend CNNs for video, fed the networks with raw frames. This makes for a much difficult learning problem. \cite{jhuang-iccv-07}, proposed a biologically inspired model for action recognition in videos with a predefined set of spatio-temporal filters in the initial layer. Combined with a similar but spatial HMAX (Hierarchical model and X) model,  \cite{kuehne2011hmdb} proposed spatial and temporal recognition streams. \cite{ji-icml-10} addressed an end-to-end learning of the CNNs for videos for the first time using $3$-D convolutions over a bunch of consecutive video frames. A more recent work by \cite{karpathy-cvpr-14}  propose a set of techniques to fuse the appearance information present from a stack of consecutive frames in a video. However, they report that the net that processes individual frames performs on par with the net that operates on a stack of frames. This might suggest that, the learnt spatio-temporal filters are not suitable to capture the motion patterns efficiently.

\begin{figure}
\centering
\includegraphics[height=7cm]{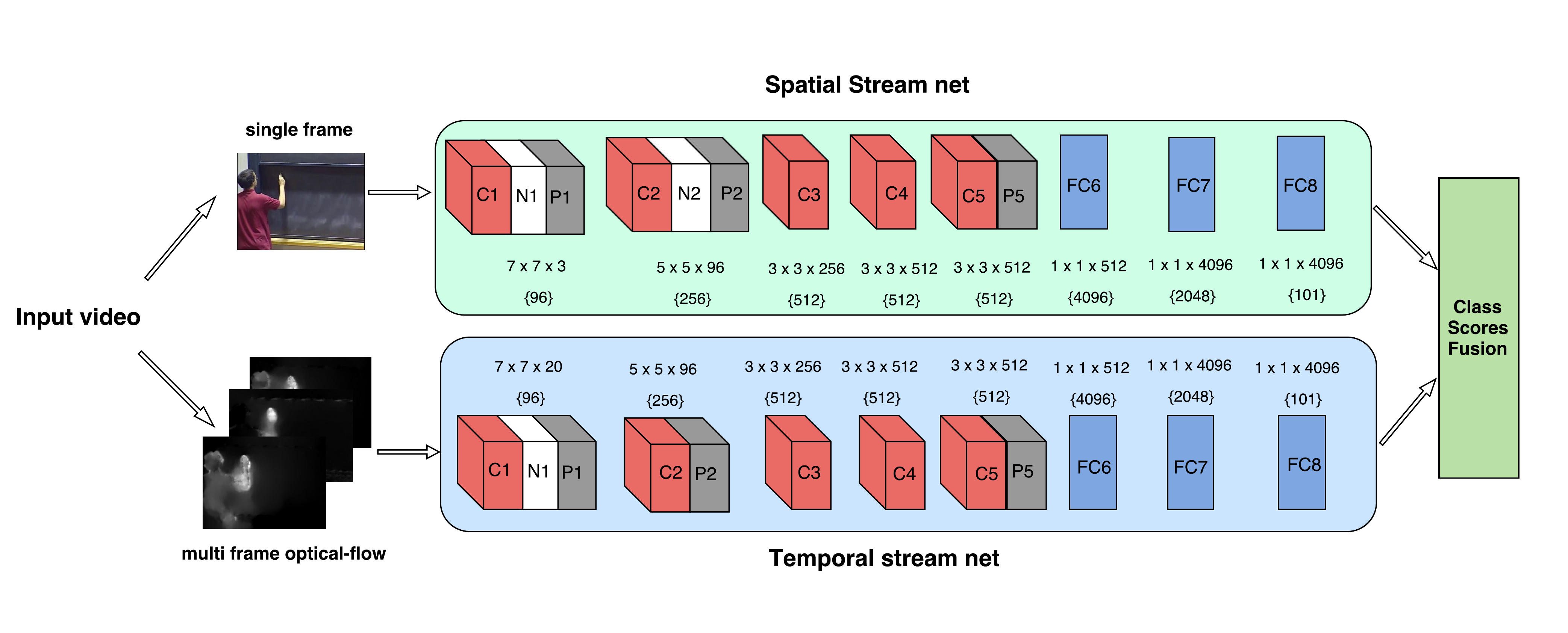}
\caption{Two stream architecture for video classification from \cite{simonyan-nips-14}}
\label{fig:TwoStreamNet}
\end{figure}

A more suitable CNN model to represent videos is proposed in a contemporaneous work by \cite{simonyan-nips-14}, which is called two-stream network approach. Though the model in \cite{kuehne2011hmdb} is also a two stream model, the main difference is that the streams are shallow and implemented with hand-crafted models. The reason for the success of this approach is the natural ability of the videos to be separated into spatial and temporal components. The spatial component in the form of frames captures the appearance information like the objects present in the video. The temporal component in the form of motion (optical flow) across the frames captures the movement of the objects. These optical flow estimates can be obtained either from classical approaches (see \cite{baker2004lucas}) or deep-learnt approaches (see \cite{weinzaepfel2013deepflow}).

This approach models the recognition system dividing into two parallel streams as depicted in Fig.\ref{fig:TwoStreamNet}. Each is implemented by a dedicated deep CNN, whose predictions are later fused. The net for the spatial stream is similar to the image recognition CNN and processes one frame at a time. However, the temporal stream takes the stacked optical flow of a bunch of consecutive frames as input and predicts the action. Both the nets are trained separately with the corresponding input. An alternative motion representation using the trajectory information similar to \cite{wang-iccv-13} is also observed to perform similar to optical flow.

The most recent methods that followed \cite{simonyan-nips-14} have similar two-stream architecture. However, their contribution is to find the most active spatio-temporal volume for the efficient video representation. Inspired from the recent progress in the object detection in images, \cite{georgia-cvpr-15} built action models from shape and motion cues. They start from the image proposals and select the motion salient subset of them and extract saptio-temporal features to represent the video using the CNNs.

\cite{wang-cvpr-15} employ deep CNNs to learn discriminative feature maps and conduct trajectory constrained pooling to summarize into an effective video descriptor. The two streams operate in parallel extracting local deep features for the volumes centered around the trajectories.

In general, these multi-modal CNNs can be modified and extended to suit any other kind of modality like audio, text to complement the image data leading to a better representation of image content.

\subsection{CNNs with RNNs}
While CNNs have made remarkable progress in various tasks, they are not very suitable for learning sequences. Learning such patterns requires memory of previous states and feedback mechanisms which are not present in CNNs. RNNs are neural nets with at least one feedback connection. This looping structure enables the RNN to have an internal memory and to learn temporal patterns in data.

\begin{figure}
\centering
\includegraphics[scale = 0.4]{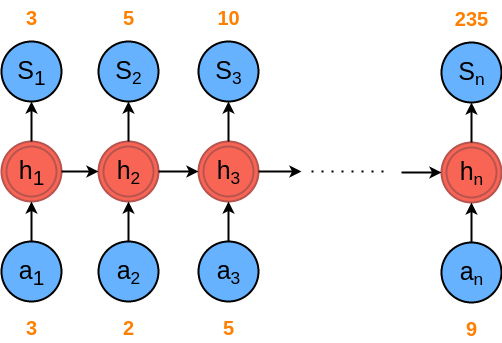}
\vspace{-0.5 cm}
\caption{Toy RNN Example: Problem of sequence addition. The inputs and outputs are shown in blue. The red cells correspond to the hidden units. An unrolled version of the RNN is shown.}
\label{fig:rnn_toy}
\end{figure}

Figure \ref{fig:rnn_toy} shows the unrolled version of a simple RNN applied to a toy example of sequence addition. The problem is defined as follows: Let $a_t$ be a positive number, corresponding to the input at time $t$. The output at time $t$ is given by 

\begin{equation*}
  S_t = \sum_{i = 1}^{t} a_i
\end{equation*}

We consider a very simple RNN with just one hidden layer. The RNN can be described by equations below.
\begin{align*}
h_{t+1} &= f_h(W_{ih} \times a_t + W_{hh} \times h_t)\\
S_{t+1} &= f_o(W_{ho} \times h_{t+1})
\label{eq:rnnEq}
\end{align*}

where $W_{ih}, W_{hh}, W_{ho}$ are learned weights and $f_h$ and $f_o$ are non-linearities. For the toy problem considered above, the weights learned would result in $W_{ih} = W_{hh} = W_{ho} = 1$. Let us consider the non-linearity to be ReLu. The equations would then become,
\begin{align*}
h_{t+1} &= ReLu(a_t +  h_t)\\
S_{t+1} &= ReLu(h_{t+1})
\end{align*}
Thus, as shown in Figure \ref{fig:rnn_toy}, the RNN stores previous inputs in memory and learns to predict the sum of the sequence up to the current timestep $t$.

As with CNNs, recurrent neural networks have been trained with various back propagation techniques. These conventional methods however, resulted in the \textit{vanishing gradient problem}, i.e. the errors sent backward over the network, either grew very large or vanished leading to problems in convergence. In 1997, \cite{hochreiter1997long} introduced LSTM (Long Short Term Memory), which succeeded in overcoming the vanishing gradient problem, by introducing a novel architecture consisting of units called Constant Error Carousels. LSTMs were thus able to learn very deep RNNs and successfully remembered important events over long (thousands of steps) durations of time.

Over the next decade, LSTM's became the network of choice for several sequence learning problems, especially in the fields of speech and handwriting recognition (see \cite{graves2009novel, graves2013speech}). In the sections that follow, we shall discuss applications of RNNs in various computer vision problems. 

\subsubsection{Action recognition}

\begin{figure}
\centering
\includegraphics[height=7cm,trim={0cm 19cm 0cm 0cm},clip]{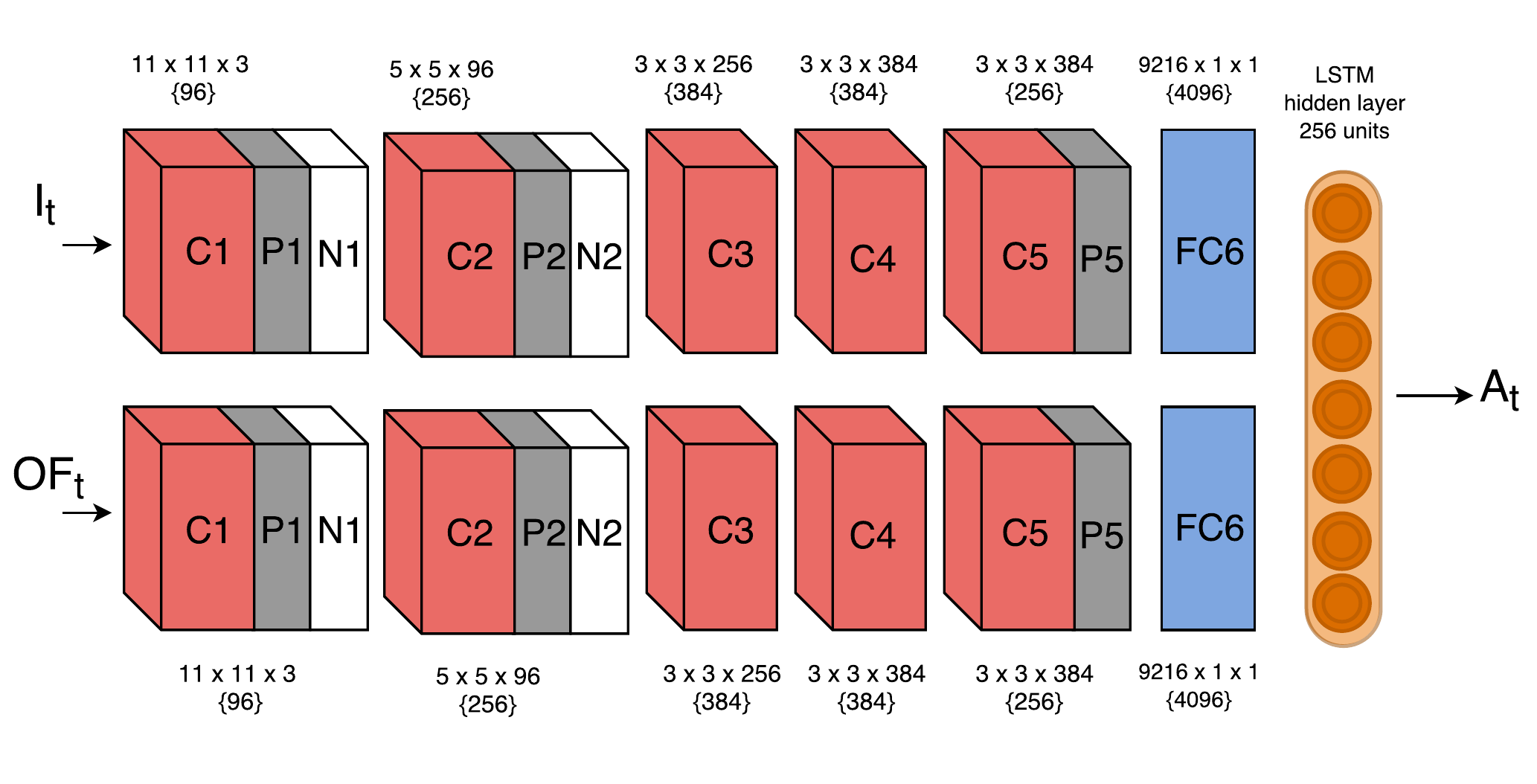}
\vspace{-0.75 cm}
\caption{LRCN: A snapshot of the model at time \textit{t}. $I_t$ refers to the frame and ${OF}_t$ the optical flow at $t$. The video is classified by averaging the output $A_t$ over all $t$.}
\label{fig:LRCN}
\end{figure}

Recognizing human actions from videos has long been a pivotal problem in the tasks of video understanding and surveillance. Actions, being events which take place over a finite length of time, are excellent candidates for a joint CNN-RNN model.  

In particular, we discuss the model proposed by \cite{donahue2014long}. They use \textit{RGB} as well as \textit{optical flow} features to jointly train a variant of Alexnet combined with a single layer of LSTM (256 hidden units). Frames of the video are sampled, passed through the trained network and classified individually. The final prediction is obtained by averaging across all the frames. A snapshot of this model at time $t$ is shown in Figure~\ref{fig:LRCN}.

\subsubsection{Image and video captioning}
Another important component of scene understanding is the textual description of images and videos. Relevant textual description also helps complement image information, as well as form useful queries for retrieval. 

RNNs (LSTMs) have long been used for machine translation (see \cite{bahdanau2014neural,cho2014learning}). This has motivated its use for the purpose of image description. \cite{vinyals2014show} have developed an end-to-end system, by first encoding an image using a CNN and then using the encoded image as an input to a language generating RNN. \cite{karpathy2014deep} propose a multimodal deep network that aligns various interesting regions of the image, represented using a CNN feature, with associated words. The learned correspondences are then used to train a bi-directional RNN. This model is able, not only to generate descriptions for images, but also to localize different segments of the sentence to their corresponding image regions. The multimodal RNN (m-RNN) by \cite{mao2014deep} combines the functionalities of the CNN and RNN by introducing a new multimodal layer, after the embedding and recurrent layers of the RNN. \cite{mao2015learning} further extend the m-RNN by incorporating a transposed weight sharing strategy, enabling the network to learn novel visual concepts from images.

\cite{venugopalan2014translating} move beyond images and obtain a mean-pooled CNN representation for a video. They train an LSTM to use this input to generate a description for the video. They further improve upon this task by developing S2VT \cite{venugopalan2015sequence} a stacked LSTM model which accounts for both the RGB as well as flow information available in videos. \cite{pan2015jointly} use both 2-D and 3-D CNNs to obtain a video embedding. They introduced two types of losses which are used to train both the LSTM and the visual semantic embedding.

\subsubsection{Visual Question answering} 
Real understanding of an image should enable a system not only to make a statement about it, but also to answer questions related to it. Therefore answering questions based on visual concepts in an image is the next natural step for machine understanding algorithms. Doing this, however, requires the system to model both the textual question and the image representation, before generating an answer conditioned on both the question and the image. 
 
A combination of CNN and LSTM has proven to be effective in this task too, as evidenced by the work of  \cite{malinowski2015ask} who train an LSTM layer to accept both the question as well a CNN representation of the image and generate the answer.   \cite{gao2015you} use two LSTM's with shared weights along with a CNN for the task. Their experiments are performed on a multilingual dataset containing Chinese questions and answers along with its English translation. \cite{antol2015vqa} provide a dataset for the task of visual question answering containing both real world images and abstract scenes.

\subsection{Hybrid learning methods}

\begin{figure}
\centering
\includegraphics[width=15cm,trim={1cm 16cm 1cm 7cm},clip]{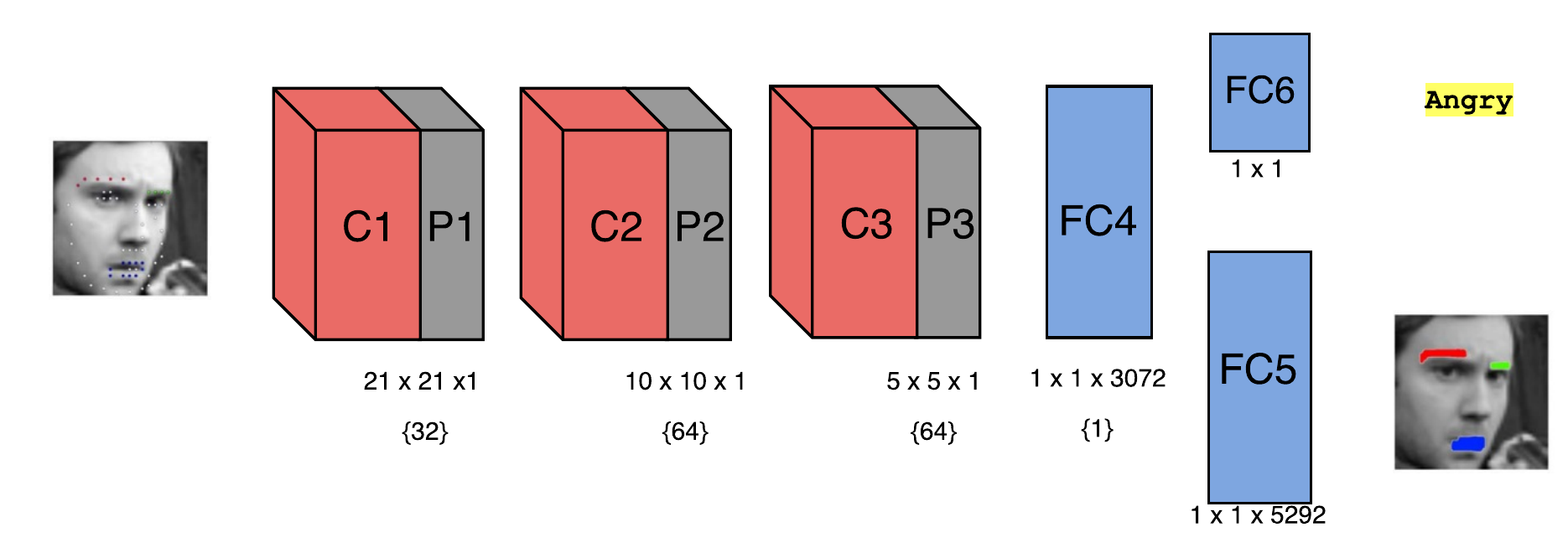}
\vspace{-0.75 cm}
\caption{The facial expression recognition system of \cite{DBLP:conf/crv/DevriesBT14} which utilizes facial landmark (shown overlaid on the face towards the right of the image) recognition as an auxiliary task which helps improve performance on the main task of expression recognition.}
\label{fig:devries}
\end{figure}

\subsubsection{Multi-task learning}
Multi-task learning is essentially a machine learning paradigm wherein the objective is to train the learning system to perform well on multiple tasks. Multi-task learning frameworks tend to exploit shared representations that exist among the tasks to obtain a better generalization performance than counterparts developed for a single task alone. 

In CNNs, multi-task learning is realized using different approaches. One class of approaches utilize a multi-task loss function with hyper-parameters typically regulating the task losses. For example, \cite{DBLP:journals/corr/Girshick15} employ a multi-task loss to train their network jointly for classification and bounding-box regression tasks thereby improving performance for object detection. \cite{DBLP:journals/corr/ZhangLLT14} propose a facial landmark detection network which adaptively weights auxiliary tasks (e.g. head pose estimation, gender classification, age estimation) to ensure that a task that is deemed not beneficial to accurate landmark detection is prevented from contributing to the network learning. \cite{DBLP:conf/crv/DevriesBT14} demonstrate improved performance for facial expression recognition task by designing a CNN for simultaneous landmark localization and facial expression recognition. A hallmark of these approaches is the division of tasks into primary task and auxiliary task(s) wherein the purpose of the latter is typically to improve the performance of the former (see Figure \ref{fig:devries}).

Some approaches tend to have significant portions of the original network modified for multiple tasks. For instance, \cite{DBLP:journals/corr/SermanetEZMFL13} replace pre-trained layers of a net originally designed to provide spatial (per-pixel) classification maps with a regression network and fine-tune the resulting net to achieve simultaneous classification, localization and detection of scene objects. 

Another class of multi-task approaches tend to have task-specific sub-networks as a characteristic feature of CNN design. \cite{humanposedcnn} utilize separate sub-networks for the joint point regression and body part detection tasks. \cite{DBLP:journals/corr/WangZLLZ15} adopt a serially stacked design wherein a localization sub-CNN and the original object image are fed into a segmentation sub-CNN to generate its object bounding box and extract its segmentation mask. To solve an unconstrained word image recognition task, \cite{DBLP:journals/corr/JaderbergSVZ14b} propose an architecture consisting of a character sequence CNN and an N-gram encoding CNN which act on an input image in parallel and whose outputs are utilized along with a CRF model to recognize the text content present within the image.

\subsubsection{Similarity learning}
Apart from classification, CNNs can also be used for tasks like metric learning and rank learning. Rather than asking the CNN to identify objects, we can instead ask it to verify whether two images contain the same object or not. In other words, we ask the CNN to learn which images are \textit{similar}, and which are not. Image retrieval is one application where such questions are routinely asked.

\indent Structurally, Siamese networks resemble two-stream networks discussed previously. However, the difference here is that both `streams' have identical weights. Siamese networks consist of two seperate (but identical) networks, where two images are fed in as input. Their activations are combined at a later layer, and the output of the network consists of a single number, or a \textit{metric}, which is a notion of distance between the images. Training is done so that images which are considered to be similar have a lower output score than images which are considered different. \cite{bromley1993signature} introduced the idea of Siamese networks and used it for signature verification. Later on, \cite{chopra2005learning} extended it for face verification. \cite{zagoruyko2015learning} further extended and generalized this to learning similarity between image patches.

\indent Triplet networks are extensions of siamese networks used for rank learning. \cite{wang2014learning} first used this idea for learning fine-grained image similarity learning. 

\section{Open problems}

In this section, we briefly mention some open research problems in deep learning, particularly of interest to computer vision. Several of these problems are already being tackled in several works. 
\begin{itemize}
\item Training CNNs requires tuning of a large number of hyper-parameters, including those involving the model architecture. An automated way of tuning such as that by \cite{snoek2012practical} is crucial for practitioners. However, that requires multiple models to be trained, which can be both time consuming and impractical for large networks.
\item  \cite{nguyen2014deep} showed that one can generate artificial images that result in CNNs producing a high confidence false prediction. In a related line of work, \cite{szegedy2013intriguing} showed that natural images can be modified in an imperceptible manner to produce a completely different classification label. Although \cite{goodfellow2014explaining} attempted to reduce the effects of such adversarial examples, it remains to be seen whether that can be completely eliminated.
\item It is well known (see \cite{gong2014multi}) that CNNs are \textit{robust} to small geometric transforms. However, we would like them to be \textit{invariant}. The study of \textit{invariance} for extreme deformation is largely missing.
\item Along with using a large number of data points, CNN models are also large and (relatively) slow to evaluate. While there has been a lot of work in reducing number of parameters (see \cite{hinton2015distilling, denil2013predicting, collins2014memory, jaderberg2014speeding, srinivas2015data}), it is not clear how to train non-redundant models in the first place.
\item CNNs are presently trained in a \textit{one-shot} way. The formulation of an \textit{online} method of training would be desirable for robotics applications.
\item Unsupervised learning is one more area where we expect to deploy deep learning models. This would enable us to leverage the massive amounts of unlabelled image data on the web. Classical deep networks like autoencoders and restricted boltzmann machines were formulated as unsupervised models. While there has been a lot of interesting recent work in the area (see \cite{goodfellow2014generative, bengio2013deep, kingma2013auto, kulkarni2015deep}), a detailed discussion of these is beyond the scope of this paper.
\end{itemize}

\section{Concluding remarks}
In this article, we have surveyed the use of deep learning networks - convolutional neural networks in particular - for computer vision. This enabled complicated hand-tuned algorithms being replaced by single monolithic algorithms trained in an end-to-end manner. However, despite our best efforts it may not be possible to capture the entire gamut of deep learning research - even for computer vision - in this paper. We point the reader to other reviews, specifically those by \cite{bengio2009learning}, \cite{lecun2015deep} and \cite{schmidhuber2015deep}. These reviews are more geared towards deep learning in general, while ours is more focussed on computer vision. We hope that our article will be useful to vision researchers beginning to work in deep learning.

\bibliography{test}
\bibliographystyle{apalike}

\end{document}